\documentclass{article}
\usepackage{ijcai18}
\usepackage{comment}

\usepackage{times}
\usepackage{xcolor}
\usepackage{soul}
\usepackage[utf8]{inputenc}
\usepackage[small]{caption}

\usepackage[notheorems]{rr-math}
\usepackage{blkarray}

\usepackage{nicefrac}

\newcommand{\cellcolor}[1]{} 

\usepackage{xcolor}

\usepackage{oubraces}

\usepackage{verbatim}
\usepackage{wrapfig}

\usepackage{booktabs}

\newcommand{\dataName}[1]{\ensuremath{\fontsize{6}{7.5}\selectfont \mathsf{#1}}} 

\usepackage{graphicx}
\usepackage{multirow}
\usepackage{rotate}
\usepackage{subfigure}
\usepackage{epstopdf}
\usepackage{paralist}
\usepackage{url}
\usepackage{graphicx}
\usepackage{wrapfig}
\usepackage{graphicx}

\usepackage{amssymb}
\usepackage{amsmath}
\usepackage{amsfonts}

\newcommand{\todo}[1]{}

\newcommand{\incomplete}[1]{}

\usepackage{tabularx}
\usepackage{booktabs}

\usepackage{relsize}
\usepackage{algorithm}
\usepackage{algorithmicx}
\usepackage{algpseudocode}

\algblockdefx[parallel]{ParFor}{EndPar}[1][]{$\textbf{parallel for}$ #1 $\textbf{do}$}{$\textbf{end parallel}$}
\algrenewcommand{\alglinenumber}[1]{\fontsize{6.5}{7}\selectfont#1}
\algtext*{EndPar}

\algblockdefx[parallel]{parfor}{endpar}[1][]{$\textbf{parallel for}$ #1 $\textbf{do}$}{$\textbf{end parallel}$}
\algrenewcommand{\alglinenumber}[1]{\scriptsize#1:}

\usepackage{nicefrac}

\algnotext{EndFor}
\algnotext{EndIf}
\algnotext{EndProcedure}

\ifpdf
\usepackage{epstopdf}
\fi
\graphicspath{{./}{./figures/}}

\graphicspath{{graphics/}{../graphics/}{./}}

\usepackage{array}
\newcolumntype{P}[1]{>{\centering\arraybackslash}p{#1}}
\newcolumntype{M}[1]{>{\centering\arraybackslash}m{#1}}

\newtheorem{thm}{Theorem}
\newtheorem{mylem}[thm]{Lemma}
\newtheorem{mydef}{Definition}

\newcommand{\eol}{\end{enumerate}\setlength{\itemsep}{-\parsep}}

\relax

\usepackage{wrapfig}
\usepackage{xspace}

\usepackage{csquotes}

\newcommand{\etal}{\emph{et al.}\xspace}

\newlength{\commentWidth}
\newcommand{\bspacing}{\begin{spacing}{1.4}}
\newcommand{\espacing}{\end{spacing}}

\definecolor{plotblue}{RGB}	{30,144,255}
\definecolor{plotgreen}{RGB}	{50,205,50}
\definecolor{plotred}{RGB}	{220,20,60}

\definecolor{myyellow}{RGB}{255,255,204}
\definecolor{myred}{RGB}{255,204,204}
\definecolor{myblue}{RGB}{204, 255, 255}
\definecolor{mygreen}{RGB}{204, 255, 204}

\definecolor{gray}{RGB}{150,150,150}
\definecolor{theblue}{RGB}{0,0,180}

\usepackage{textcase}
\usepackage{soul}

\newcommand*\hrulefillvar[1][0.4pt]{\leavevmode\leaders\hrule height#1\hfill\kern0pt}

\newcommand{\Alpha}{A}

\newcommand{\be}{\begin{equation}}
\newcommand{\ee}{\end{equation}}
\newcommand{\bea}{\begin{eqnarray}}
\newcommand{\eea}{\end{eqnarray}}
\newcommand{\bit}{\begin{itemize}}
\newcommand{\eit}{\end{itemize}}

\definecolor{lightgray}{rgb}{0.93,0.93,0.93}

\definecolor{lightblue}{rgb}{0.5,0.90,1.0}
\definecolor{lightgreen}{rgb}{0.5,0.92,0.5}
\definecolor{lightred}{rgb}{0.98,0.5,0.5}
\definecolor{lightyellow}{rgb}{1,0.90,0.40}

\newcommand{\verylightgreen}{green!10} 

\renewcommand{\verylightgreen}{blue!10}

\usepackage{booktabs}
\usepackage{comment}

\newtheorem{property}{Property}[section]  

\definecolor{myyellow}{RGB}{255,255,204}
\definecolor{myred}{RGB}{255,204,204}
\definecolor{myblue}{RGB}{0,200,255}
\definecolor{mygreen}{RGB}{80,220,80}

\newcommand{\eg}{\emph{e.g.}}
\newcommand{\ie}{\emph{i.e.}}

\algblockdefx[parallel]{parallelfor}{parallelend}
[1][]{\textbf{parallel for} #1}
{\textbf{end parallel}}

\RequirePackage{mathtools}
\RequirePackage{color}
\RequirePackage{listings}

\newcommand{\ds}\displaystyle
\newcommand{\mbb}\mathbb
\newcommand{\mc}\mathcal
\newcommand{\del}\nabla

\newcommand{\beqstar}{\begin{eqnarray*}}
\newcommand{\eeqstar}{\end{eqnarray*}}

\definecolor{thegreen}{rgb}{0,.5,0}
\definecolor{idea}{rgb}{0,.6,0.1}
\definecolor{problem}{rgb}{0.7,0,0.1}
\definecolor{comment-green}{rgb}{0,.3,0}
\definecolor{theblue}{rgb}{0,0,.8}
\definecolor{light-gray}{gray}{0.98}
\definecolor{comment-color}{rgb}{0,0,.8}
\definecolor{string-color}{rgb}{0,.75,0}
\definecolor{border-blue}{rgb}{0,0,.6}

\usepackage{epsfig}

\newcommand{\ra}[1]{\renewcommand{\arraystretch}{#1}} 

\usepackage{ragged2e}

\usepackage{setspace}

\graphicspath{{./}{./images/}}
\newcolumntype{H}{>{\setbox0=\hbox\bgroup}c<{\egroup}@{}}

\definecolor{orange}{rgb}{1,0.5,0}
\definecolor{gray}{RGB}{20,20,20}
\definecolor{greencm}{RGB}{0,153,0}

\definecolor{gray}{RGB}{150,150,150}
\definecolor{theblue}{RGB}{0, 20, 159}

\newcommand{\href}[2]{\ttfamily\upshape
$\mathtt{#1}$}

\usepackage{ifpdf}
\usepackage{paralist}
\usepackage{tabularx}
\usepackage{booktabs}
\usepackage{multirow}
\usepackage{graphicx}
\usepackage{sidecap}
\usepackage{mathdots}
\usepackage{lscape}
\makeatletter
\def\vcdots{\vbox{\baselineskip4\p@ \lineskiplimit\z@
\kern3\p@\hbox{.}\hbox{.}\hbox{.}\kern3\p@}}
\makeatother

\usepackage{subfigure}
\usepackage{array}
\usepackage{url}
\newcommand{\ourMethod}{\ensuremath{\mathrm{\text{our method}}}}

\newcommand{\deepwalk}{\textrm{DeepWalk}}

\providecommand{\Y}{\ensuremath{W}} 
\providecommand{\y}{\ensuremath{w}}

\providecommand{\RR}{\mathbb{R}}

\providecommand{\Std}{\mathbb{S}}

\providecommand{\eproof}{$\null\hfill\square$}

\renewcommand{\phi}{\ensuremath{\Phi}}

\providecommand{\corr}{\ensuremath{\mathbb{K}}}
\providecommand{\sim}{\ensuremath{\corr}}

\providecommand{\E}{\ensuremath{E}}

\providecommand{\p}{\ensuremath{p}}

\providecommand{\x}{\ensuremath{\sx}}  
\providecommand{\y}{\ensuremath{\sy}}

\providecommand{\p}{\ensuremath{p}}

\renewcommand{\O}{\ensuremath{\mathcal{O}}}
\renewcommand{\bigO}[1]{\ensuremath{\mathcal{O}(#1)}}

\providecommand{\m}{\ensuremath{M}} 
\providecommand{\n}{\ensuremath{N}}

\renewcommand{\k}{\ensuremath{K}}  

\providecommand{\nL}{\ensuremath{L}}  
\providecommand{\nR}{\ensuremath{R}}  

\providecommand{\nD}{\ensuremath{D}}

\providecommand{\N}{\ensuremath{\Gamma}}

\title{Learning Role-based Graph Embeddings}

\def\fontsz{}

\newcommand{\authorEmail}[1]{}

\author{
Nesreen K. Ahmed\\
Intel Labs\\
\authorEmail{nesreen.k.ahmed@intel.com}
\And
Ryan A. Rossi\\
Adobe Research\\
\authorEmail{rrossi@adobe.com}
\And
John Boaz Lee\\
\fontsz WPI\\
\authorEmail{jtlee@wpi.edu}
\AND
\fontsz
Ted Willke\\
Intel Labs \\
\authorEmail{ted.willke@intel.com}
\And
Rong Zhou\\
\fontsz Google\\
\authorEmail{rongzhou@google.com}
\And
Xiangnan Kong\\
\fontsz WPI\\
\authorEmail{xkong@wpi.edu}
\And
Hoda Eldardiry\\
\fontsz Xerox PARC\\
\authorEmail{heldardiry@parc.com}
}

\begin{document}

\maketitle

\begin{abstract}
Random walks are at the heart of many existing network embedding methods.
However, such algorithms have many limitations that arise from the use of random walks, \eg, the features resulting from these methods are unable to transfer to new nodes and graphs as they are tied to vertex identity.
In this work, we introduce the \emph{Role2Vec} framework which uses the flexible notion of \emph{attributed random walks}, and serves as a basis for generalizing existing methods
such as DeepWalk, node2vec, and many others that leverage random walks.
Our proposed framework enables these methods to be more widely applicable for both transductive and inductive learning as well as for use on graphs with attributes (if available). This is achieved by learning functions that generalize to new nodes and graphs.
We show that our proposed framework is effective with an average AUC improvement of $16.55\%$ while requiring on average 853x less space than existing methods on a variety of graphs.
\end{abstract}

\section{Introduction}
\label{sec:intro}
Learning a useful feature representation from graph data lies at the heart and success of many machine learning tasks such as node classification~\cite{neville2000iterative}, anomaly detection~\cite{akoglu2015graph}, link prediction~\cite{al2011survey}, among others~\cite{koyuturk2006pairwise,ng2002spectral}. Motivated by the success of word embedding models, such as the skip-gram model~\cite{mikolov2013distributed}, recent works extended word embedding models to learn graph embeddings~\cite{deepwalk,goyal2017graph}. The primary goal of these works is to model the conditional probabilities that relate each input vertex to its context, where the context is a set of other vertices surrounding and/or topologically related to the input vertex. Many variants of graph embedding methods proposed \emph{random walks} to generate the context vertices~\cite{deepwalk,node2vec,struc2vec,ComE}. For instance, DeepWalk~\cite{deepwalk} initiates random walks from each vertex to collect sequences of vertices (similar to sentences in language). Then, the skip-gram model is used to fit the embeddings by maximizing the conditional probabilities that relate each input vertex to its surrounding context. In this case, vertex identities are used as words in the skip-gram model, and the embeddings are tied to the vertex ids.

In language, the foundational idea is that words with similar meanings will be surrounded by a similar context~\cite{harris1954distributional}. As such, in language models, the context of a word is defined as the surrounding words. However, this foundation does not directly translate to graphs. Since unlike words in languages that are universal with semantics and meaning independent of the corpus of documents, vertex ids obtained by random walks on graphs are not universal and are only meaningful within a particular graph. This key limitation has two main disadvantages. First, these embedding methods are inherently transductive, dealing essentially with isolated graphs, and unable to generalize to unseen nodes. Consequently, they are unsuitable for graph-based transfer learning tasks such as across-network classification~\cite{kuwadekar2011relational,introSRL07}, and graph similarity/comparison~\cite{goldsmith1990assessing,zager2008graph}. Second, by using this traditional definition of random walks, there is no general way to integrate vertex attributes/features to the network representation.

There is no guarantee that similar vertices are surrounded by similar context (obtained using random walks on graphs). Recent empirical analysis shows that using random walks in graph embeddings primarily capture proximity among the vertices (see~\cite{goyal2017graph}), so that vertices that are close to one another in the graph are embedded together, \eg, vertices that belong to the same community are embedded similarly. Although proximity among the vertices does not guarantee their similarity, the idea of a network position or a \emph{role}~\cite{lorrain1977structural,rossi2014roles,henderson2011s} is more suitable to represent the similarity and structural relatedness among vertices. Roles represent vertex connectivity patterns such as hubs, star-centers, star-edge nodes, near-cliques or vertices that act as bridges to different regions of the graph.  Intuitively, two vertices belong to the same role if they are structurally similar. Random walks will likely visit nearby vertices first, which makes them suitable for finding communities, rather than roles (structural similarity) (see Sec.~\ref{sec:analysis} for theoretical analysis). 

To overcome the above problems, we propose the \emph{Role2Vec} framework which serves as a basis for generalizing many existing methods that use traditional random walks.
\emph{Role2Vec} utilizes the flexible notion of attributed random walks that is not tied to vertex identity and is instead based on a function $\phi : \vx \rightarrow \y$ that maps a vertex attribute vector to a type, such that two vertices belong to the same type if they are structurally similar.
The proposed framework provides a number of important advantages to any method generalized using it.
First, the proposed framework is 
naturally inductive as the learned features generalize to new nodes and across graphs and therefore can be used for transfer learning tasks.
Second, they are able to capture structural similarity (roles) better.
Third, the proposed framework is inherently space-efficient since embeddings are learned for types (as opposed to vertices) and therefore requires significantly less space than existing methods.
Fourth, the proposed framework naturally supports graphs with attributes (if available/given as input).
Furthermore, our approach is shown to be effective with an average improvement of $16.55\%$ in AUC while requiring on average $853$x less space than existing methods on a variety of graphs from different application domains.

\section{Framework} 
\label{sec:framework}

\setlength{\abovedisplayskip}{3pt}
\setlength{\belowdisplayskip}{3pt}

We consider an (un)directed input graph $G=(V,E)$, where $N_v = |V|$ is the number of vertices in $G$, and $N_e = |E|$ is the number of edges in $G$.  For any vertex $v_i \in V$, let $\Gamma(i)$ be the set of direct neighbors of $v_i$, and $d_i = |\Gamma(i)|$ is the vertex degree. In addition, we consider a matrix $\mX$ of attributes/features, where each $\vx_i$ is a $K$-vector for vertex $v_i$. For example, for graphs without attributes, $\vx_i$ could simply be an indicator vector for vertex $v_i$ and $K$ is equivalent to the number of vertices (\ie, having $x_{ij} =1$ if $j=i$, and $x_{ij} =0$ otherwise)~\cite{deepwalk,node2vec}.  For attributed graphs, $\vx_i$ may include observed attributes, topological features, and/or node types for heterogeneous graphs. 
The goal of an embedding method is to derive useful features of particular graph elements (\eg, vertices, edges) by learning a model that maps each graph element to the latent $D$-dimension space. While the approach remains general for any graph element, this paper focuses on vertex embeddings. 

To achieve this, an embedding is usually defined with three components: (1) the context function, which specifies a set of other vertices called the \emph{context} $c_i$ for any given vertex $v_i$, such that the context vertices are surrounding and/or topologically related to the given vertex. Each vertex is associated with two latent vectors, an \emph{embedding} vector $\boldsymbol\alpha_i \in \mathbb{R}^D$ and a \emph{context} vector $\boldsymbol\beta_i \in \mathbb{R}^D$. (2) the conditional distribution, which specifies the statistical distribution used to combine the embedding and context vectors. More specifically, the conditional distribution of a vertex combines its embedding and the context vectors of its surrounding vertices. (3) the model parameters (\ie, embedding and context vectors) and how these are shared across the conditional distributions. Thus, an embedding method models the conditional probability that relate each vertex to its context as follows: $\vx_{c_i}\;|\;\vx_i \sim \mathbb{P}$,  
where $c_i$ is the set of context vertices for vertex $v_i$, $\vx_i$ is its feature/attribute vector, and $\mathbb{P}$ is the conditional distribution. 

Our goal is to model $\mathbb{P}[\vx_{c_i} | \vx_i] = \prod_{j \in c_i} \mathbb{P}(\vx_j | \vx_i)$, assuming the context vertices are conditionally independent. The most commonly used conditional distribution is the categorical distribution (see~\cite{rudolph2016exponential} for a generalization). In this case, a softmax function parameterized with the two latent vectors (\ie, embedding and context vectors) is used. Thus, for each input-context vertex pair $(v_i,v_j)$,
{\small
\begin{align}
\mathbb{P}(\vx_j | \vx_i) = \frac{\operatorname{e}^{\boldsymbol\alpha_i . \boldsymbol\beta_j}}{\sum_{v_k \in V} \operatorname{e}^{\boldsymbol\alpha_i . \boldsymbol\beta_k}}
\end{align}\noindent
\normalsize}
For sparse graphs, the summation in $\sum_{v_k \in V} \operatorname{e}^{\boldsymbol\alpha_i . \boldsymbol\beta_k}$ contains many zero entries, and thus can be approximated by sub-sampling those zero entries (using negative sampling similar to language models~\cite{mikolov2013distributed}). Finally, the objective function of the embedding method is the sum of the logarithm of likelihood values of each vertex, \ie, {\small$\mathcal{L}(\boldsymbol\alpha,\boldsymbol\beta) = \sum_{i = 1}^{N_v} \log\;\mathbb{P}[\vx_{c_i} | \vx_i]$\normalsize}.

Clearly, there is a class of possible embedding methods where each of the three components (discussed above) is considered a modeling choice with various alternatives. Recent work proposed \emph{random walks} to sample/collect the context vertices $c_i$~\cite{deepwalk,node2vec}. 

\subsection{Mapping Vertices to Vertex-Types}
\label{sec:framework-function-mapping-nodes-to-types}
Given $N_v \times \k$ matrix $\mX$ of attributes and/or structural features, the \emph{Role2Vec} framework starts by locating sets of vertices, however large or small be the shortest distance between any two in a set, who are placed similarly with respect to all other sets of vertices. Thus, two vertices belong to the same set if they are similar in terms of attributes and/or structural features. We achieve this by learning a function that maps the $N_v$ vertices to a set $\Y = \{ \y_{1},...,\y_{\m} \}$ of $\m$ \emph{vertex-types} 
where $\m$ is often much smaller than $N_v$, \ie, $\m \ll N_v$,
{\small
\begin{gather}
\phi \; : \; \vx \; \rightarrow \; \y
\end{gather}
\normalsize}
\noindent
Thus, $\phi$ is a function mapping vertices to \emph{vertex-types} based on the $N_v \times \k$ attribute matrix $\mX$. Clearly, the function $\phi$ is a modeling choice, which could be learned automatically or defined manually by the user. We explore two general classes of functions for mapping vertices to their types.
The first class of functions are simple functions taking the form:
{\small
\begin{gather}\label{eq:simple-function}
\phi(\vx) = x_1 \circ x_2 \circ \cdots \circ x_{\k}
\end{gather} 
\normalsize}
\noindent
where $\vx = \big[\; x_1\;\, x_2\;\, \cdots \;\, x_{\k} \; \big]$ is an attribute vector and $\circ$ is a 
binary operator such as concatenation, sum, among others.
The second class of functions are learned by solving an objective function.
This includes functions based on a low-rank factorization of the $N_v \times \k$ matrix $\mX$ having the form $\mX \approx f\langle\mU\mV^T\rangle$ with factor matrices $\mU \in \RR^{N_v \times r}$ and $\mV \in \RR^{\k \times r}$ where $r>0$ is the rank and $f$ is a linear or non-linear function. 
More formally,
{\small
\begin{align}\label{eq:low-rank}
\arg\min_{\mU,\mV \in \mathcal{C}} \; 
\Big[
\mathcal{D} \big(\mX, f\langle\mU\mV^T\rangle \big) + \mathcal{R}(\mU, \mV) 
\Big]
\end{align}
\normalsize}
\noindent
where $\mathcal{D}$ is the loss, $\mathcal{C}$ is constraints (\eg, non-negativity constraints $\mU \geq 0, \mV \geq 0$), and $\mathcal{R}(\mU, \mV)$ is a regularization penalty.
Then, we partition $\mU \in \RR^{N_v \times r}$ into $\m$ disjoint sets of nodes (for each of the $\m$ vertex-types) $V_1, \ldots, V_{\m}$, where $V_j$  is set of vertices mapped to vertex-type $w_j \in W$, by solving the k-means objective:
{\small
\begin{align}\label{eq:k-means-objective-func}
\min_{\{V_j\}^{\m}_{j=1}} \sum_{j=1}^{\m} \sum_{\vu_i \in V_j} \| \vu_i - \vc_j \|^2,
\text{where } \vc_j = \frac{\sum_{\vu_i \in V_j} \vu_i}{|V_j|}
\end{align}
\normalsize}
\vspace{-2mm}

\subsection{Attributed Random Walks}
\label{sec:framework-attributed-random-walks}
Recently, random walks received much attention in learning network embeddings~\cite{deepwalk,node2vec}, in particular to generate the context vertices. 
Consider a random walk of length $L$ and starting at a vertex $v_0$ of the input graph $G$, if at time $t$ we are at vertex $v_t$, then at time $t+1$, we move to a neighbor of $v_t$ with probability $1/d_{v_t}$. Thus, the resulting randomly chosen sequence of vertex indices $(v_t : t = 0, 1, ..., L-1)$ is a Markov chain. However, a key limitation of these methods is that the embeddings learned based on random walks are fundamentally tied to vertex ids. By using this traditional definition of random walks, there is no general way to integrate vertex attributes and structural features to the network representation. On the other hand, vertex attributes and structural features can easily be represented by differentiating the edges according to the types of their endpoints, which leads to the definition of \emph{attributed random walks}.

\begin{mydef}[Attributed walk]\label{def:attr-walk}
Let $\vx_i$ be a $\k$-dimensional vector for vertex $v_i$.
An \emph{attributed walk} of length $\nL$ is a sequence of adjacent vertex-types, 
{\small
\begin{gather} \label{eq:attr-random-walk}
\phi(\vx_{v_{0}}),\ldots, \phi(\vx_{v_{t}}), \phi(\vx_{v_{t+1}}),\ldots,\phi(\vx_{v_{\nL-1}})
\end{gather}
\normalsize}
induced by a randomly chosen sequence of indices $(v_t : t = 0, 1, ..., \nL-1)$ generated by a random walk of length $\nL$ starting at $v_0$, and a function $\phi : \vx \rightarrow \y$ that maps an input vector $\vx$ to a vertex type $\phi(\vx)$. 
\end{mydef}
The induced vertex-type sequence in the above definition is called \emph{attributed random walks} and is also a Markov chain.

The \emph{Role2vec} framework uses vertex mapping and attributed random walks to learn the embeddings. Thus, our goal is to model the conditional probability that relate each vertex-type to the types of its context,
{\small
\begin{align}
\mathbb{P}\Big[ \Phi\langle \vx_{c_i} \rangle | \Phi\langle \vx_i\rangle \Big] = \prod_{j \in c_i} \mathbb{P}(\Phi\langle \vx_j \rangle \ | \Phi\langle \vx_i\rangle)
\end{align}
\normalsize}
\noindent
Hence, the embedding structure (\ie, the embedding and context vectors) is shared among the vertices with the same vertex-type. Specifically, we learn $\boldsymbol\alpha_j \in \mathbb{R}^D$ and $\boldsymbol\beta_j \in \mathbb{R}^D$ for each partition $V_j$ of vertices, which are mapped to vertex-type $\y_j$. Note that \emph{Role2vec} learns an embedding for an aggregated network, where detailed relations among individual vertices are aggregated to total relations among vertex-types.
\subsection{Role2Vec Algorithm}
\label{sec:role2vec}
{\algrenewcommand{\alglinenumber}[1]{\fontsize{6.5}{7}\selectfont#1 }
\newcommand{\multiline}[1]{\State \parbox[t]{\dimexpr\linewidth-\algorithmicindent}{#1\strut}}
\begin{figure}[t!]
\vspace{-2mm}
\centering
\begin{algorithm}[H]
\caption{\,\small Role2Vec
\vspace{0.2mm}
}
\label{alg:generalized-node2vec}
{
\begin{spacing}{1.2}
\fontsize{7}{8}\selectfont
\begin{algorithmic}[1]
\Procedure{Role2Vec}{
$G = (V,\E)$ and $\mX$,
embedding dimensions $\nD$, 
walks per node $\nR$, walk length $\nL$, 
context (window) size $\omega$
}
\smallskip

\State Initialize set of \emph{attributed walks} $\mathcal{S}$ to $\emptyset$ \label{algline:node2vec-gen-init-attr-walks}

\State Extract (motif) features if needed and append to $\mX$ \label{algline:node2vec-gen-extract-features}
\State Transform each attribute in $\mX$ (\eg, using logarithmic binning) \label{algline:nod2vec-gen-transform-features}
\State Map vertices to types function $\phi \; : \; \vx \; \rightarrow \; \y$  \label{algline:map-to-type}

\State Precompute transition probabilities $\pi$ 
\label{algline:node2vec-gen-preprocess-mod-weights}

\State $G^{\prime} = (V,E,\pi)$ 

\parfor[$j = 1,2,...,\nR$] \label{algline:node2vec-gen-for-each-random-walk} \Comment{walks per node} 
	\State Set $\Pi$ to be a random permutation of the nodes $V$
	
	\For{{\bf each} $v \in \Pi$ in order} 
	
		\State $S = \textsc{AttributedWalk}(G^{\prime}, \mX, v, \Phi, \nL)$ 
		\label{algline:node2vec-gen-obtain-attr-walk}
		\State Add the \emph{attributed walk} $S$ to $\mathcal{S}$ \label{algline:node2vec-gen-add-attr-walk-to-set}
	\EndFor
\endpar \label{algline:node2vec-gen-endloop}
\State $\boldsymbol\Alpha = \textsc{StochasticGradientDescent}(\omega, \nD, \mathcal{S})$ \label{algline:node2vec-gen-SGD-with-attr-walks} 
\Comment{$\mathbf{parallel}$}
\State \textbf{return} the learned \emph{type} embeddings $\boldsymbol\Alpha$ \label{algline:node2vec-gen-return-learned-representation-matrix} 
\EndProcedure
\vspace{1mm}
\hrule
\vspace{1mm}
\Procedure{AttributedWalk}{$G^{\prime}$, $\mX$, start node $s$, function $\Phi$,
$\nL$}
\State Initialize attributed walk $S$ to $\big[ \Phi(\vx_s) \big]$ \label{algline:node2vec-gen-attr-walk-init-walk-and-add-start-node-function}
\State Set $i = s$ \Comment{current node} \label{algline:node2vec-gen-attr-walk-set-curr-node-to-start-node}
\For{$\ell = 1$ {\bf to} $\nL-1$} \label{algline:node2vec-gen-attr-walk-for}
	\State $\N_i = $ Set of the neighbors for node $i$ \label{algline:node2vec-gen-attr-walk-get-neighbors}
	\State $j = \textsc{AliasSample}(\N_i, \pi)$ \Comment{select node $j \in \N_i$} \label{algline:node2vec-gen-attr-walk-alias-sample}
	\State Append $\Phi(\vx_j)$ to $S$ \label{algline:node2vec-gen-attr-walk-add-node-function-to-list}
	\State Set $i$ to be the current node $j$ 
	\label{algline:node2vec-gen-attr-walk-curr-node-i}
\EndFor \label{algline:node2vec-gen-attr-walk-for-end}
\State \textbf{return} attributed walk $S$ of length $\nL$ rooted at node $s$ \label{algline:node2vec-gen-attr-walk-return-attr-walk}
\EndProcedure
\end{algorithmic}
\end{spacing}}
\end{algorithm}
\vspace{-4mm}
\end{figure}}

The \emph{Role2Vec} algorithm is shown in Alg.~\ref{alg:generalized-node2vec}. Alg.~\ref{alg:generalized-node2vec} takes the following inputs: (1) graph $G$, (2) attribute matrix $\mX$, (3) embedding dimension $D$, (4) walks per vertex $\nR$, (5) walk length $\nL$, (6) context window size $\omega$. In Line~\ref{algline:node2vec-gen-extract-features}, if $\mX$ is not available, we derive structural features using the graph structure itself. For instance, in this paper, we use small subgraphs called motifs as structural features. Counts of motif patterns were shown to be useful for a variety of network analysis tasks and can be computed quickly and efficiently with parallel algorithms~\cite{pgd,ahmed2016kais,benson2016higher}. Since many graph properties including motifs exhibit power law distributions, we preprocess $\mX$ using logarithmic binning, similar to~\cite{henderson2011s} (Line~\ref{algline:nod2vec-gen-transform-features}). In Line~\ref{algline:map-to-type}, vertices are mapped to vertex-types using a function $\Phi(\vx)$ as discussed in Section~\ref{sec:framework-function-mapping-nodes-to-types}. Then, we precompute the random walk transition probabilities $\pi$, which could be uniform or weighted (Line~\ref{algline:node2vec-gen-preprocess-mod-weights}). Lines~\ref{algline:node2vec-gen-for-each-random-walk}-\ref{algline:node2vec-gen-endloop} initiate random walks from each vertex using the notion of attributed random walks in Lines~\ref{algline:node2vec-gen-attr-walk-init-walk-and-add-start-node-function}--\ref{algline:node2vec-gen-attr-walk-return-attr-walk}. Finally, \emph{Role2Vec} learns the embeddings using stochastic gradient descent in Line~\ref{algline:node2vec-gen-SGD-with-attr-walks}.

Recall that $N_v$ is the number of nodes, $\m$ is the number of types, and $\m \ll N_v$. \emph{Role2vec} has the following properties.
\begin{property} \label{prop-role2vec1}
\emph{Role2vec} is space-efficient with space complexity {\small$\O(\m\nD + N_v)$\normalsize}.
\end{property}
\noindent\textsc{Proof.} To store the learned embeddings of the vertex-types, \emph{Role2vec} takes {\small$\O(\m\nD)$\normalsize} space. Also, \emph{Role2vec} takes $\O(\n)$ space for a hash table mapping vertices to their corresponding types. Thus, the total space used by \emph{Role2vec} is {\small$\O(\m\nD + \n)$\normalsize}, less space compared to baselines that require {\small$\O(N_v \nD)$\normalsize}. 
\eproof

As $\m \rightarrow N_v$, \emph{Role2vec} converges to the baseline random walk methods~\cite{deepwalk,node2vec}, since each vertex is mapped to a new type that uniquely identifies it from other vertices, \ie, $\phi$ is a one-to-one function from $V$ onto itself.

\section{Theoretical Analysis} \label{sec:analysis} 
\vspace{-1mm}
In a graph $G$, the sequence of vertices visited by a random walk of length $L$ is represented by a directed path on the graph. In this section, we analyze the properties and parameters of random walks that affect the embedding methods. Lemmas~\ref{prop1}--\ref{prop3} analyze the constraints and bounds on vertex reachability, expected access time, and representation of vertices/edges in random walks respectively.

We consider a random walk of length $L$ and starting at vertex $v_0$ of $G$, if at time $t$ we are at vertex $v_t$, then at time $t+1$, we move to a neighbor of $v_t$ with probability $1/d_{v_t}$. Clearly, the randomly chosen sequence of vertex indices $(v_t : t = 0, 1, ..., L-1)$ is a Markov chain. We denote by $P_t$ the distribution of $v_t$, where {\small$P_i(t) = \text{Pr}(v_t = i)$\normalsize} is the probability that the random walk visits vertex $i$ at time $t$. Similarly, we denote by $P_{ij}$ the transition probability from vertex $i$ to vertex $j$ in one step, where {\small$P_{ij} = \text{Pr}(v_t = j | v_{t-1} = i)$\normalsize}. Thus, the Markov property implies that this Markov chain is uniquely defined by its \emph{one-step} transition matrix {\small$\mP = (P_{ij})_{v_i, v_j \in V}$\normalsize}, 
{\small
\begin{align}
P_{ij}= 
\begin{cases}
1/d_i,& \text{if } (i,j) \in E\\
0,              & \text{otherwise}
\end{cases}
\end{align}
\normalsize}\noindent
Let $\mP^m$ be the transition matrix whose entries are the $m$-step transition probabilities, such that
{\small
\begin{align}
P_{ij}^m = \text{Pr} (v_{t+m} = j\;|\;v_t = i) 
\end{align}
\normalsize}\noindent
is the probability that the walk moves from vertex $i$ to vertex $j$ in exactly $m$ steps. 
Finally, we denote by $r_{ij}^t$ the probability that starting at vertex $i$, the \emph{first transition} to vertex $j$ occurs at time $t$,
{\small
\begin{align}
r_{ij}^t = \text{Pr} (v_t = j, \forall 1 \leq m \leq t-1, v_m \neq j\;|\;v_0 = i) 
\end{align}
\normalsize}\noindent

\begin{mylem} \label{prop1}
If $u$ and $v$ are two non-adjacent vertices in a connected graph $G$, then there is at least one neighbor $j \in \Gamma(u)$ where $r_{uv}^t \leq r_{jv}^{t_1}$ for $t_1 < t$.
\end{mylem}
\noindent\textsc{Proof.}  Let $d_u = |\Gamma(u)|$ be the degree of vertex $u$, and denote by $[d_u]$ the set of neighbors of $u$. For each neighbor $j \in [d_u]$, start a random walk at $j$, and let  $r_{jv}^{t_1}$ be the probability that the first transition from $j$ to $v$ occurs at time $t_1$. 
Now begin a random walk at $u$ and let  $r_{uv}^{t}$ be the probability that the first transition from $u$ to $v$ occurs at time $t$. By conditioning on the first transition, we have
{\small
\begin{align}
r_{uv}^{t} &= \sum_{j =1}^{d_u} P_{uj}\;.\;r_{jv}^{t-1} = \frac{1}{d_u} \sum_{j =1}^{d_u} r_{jv}^{t-1} \nonumber
\end{align}
\normalsize}\noindent
Set $t_1 = t-1$, thus the probability $r_{uv}^{t}$ is the mean of the probabilities of $u$'s neighbors, $r_{jv}^{t_1}$ for $j \in [d_u]$ and $t_1 < t$. This implies that there is at least one neighbor $j$ where $r_{uv}^t \leq r_{jv}^{t_1}$ for $t_1 < t$, and Property~\ref{prop1} is proved. \eproof

Lemma~\ref{prop1} shows that the probability $r_{uv}^t$ is upper bounded by the probability of at least one of $u$'s neighbors (\ie, $r_{jv}^{t_1}$).  
\begin{mylem} \label{prop2}
If $u$ and $v$ are two non-adjacent vertices in a connected graph $G$, with $h_{uv}$ is the expected access time from $u$ to $v$, and $\tilde{h}_{jv}$ is the average neighbor access time for $u$, then with probability less than $1/2$, a random walk starting at $u$ takes at least $L = 2\;h_{uv} > 2\;\tilde{h}_{jv}$ time to reach $v$.
\end{mylem}
\noindent\textsc{Proof.} Recall that $r_{uv}^{t}$ is the probability that starting at $u$, the random walk first visits $v$ at time $t$, then the expected access time from $u$ to $v$ is $\mathbb{E}[t] = h_{uv} = \sum_{t \geq 1} t\;.\;r_{uv}^{t}$. By conditioning on the first transition, we have  
{\small
\begin{align}
h_{uv} & = \sum_{t \geq 1} t\;.\; \sum_{j = 1}^{d_u} P_{uj}\;.\;r_{jv}^{t-1} \nonumber \\
& = \sum_{j = 1}^{d_u} P_{uj} +  \sum_{t >1}  t\;.\; \sum_{j = 1}^{d_u} P_{uj} \;.\; r_{jv}^{t} \nonumber \\
& = 1 +  \frac{1}{d_u}  \sum_{j = 1}^{d_u} \sum_{t >1}  t\;.\; r_{jv}^{t} 
= 1 +  \frac{1}{d_u} \sum_{j = 1}^{d_u} h_{jv}  \nonumber
\normalsize
\end{align}
\normalsize}\noindent
where $d_u$ is the degree of $u$, and $h_{jv}$ is the expected access time for some neighbor vertex $j \in [d_u]$.
Since $t \geq 0$ for any vertex in $G$, then by Markov's inequality, for any $L > 0$, 
{\small
\begin{align}
\text{Pr} (t \geq L) &\leq \frac{\mathbb{E}[t]}{L} = \frac{h_{uv}}{L} \nonumber
\end{align}
\normalsize}\noindent
Let $\tilde{h}_{jv} =  \frac{1}{d_u} \sum_{j = 1}^{d_u} h_{jv}$ be the average neighbor access time for vertex $u$. Then, with $\text{Pr} (t \geq L) \leq 1/2$, a random walk starting at $u$ takes at least $L = 2\;h_{uv} > 2\;\tilde{h}_{jv}$ to reach $v$. \eproof

Lemma~\ref{prop2} shows that the expected access time for a random walk from $u$ to $v$ is at least twice the average neighbor access time for vertex $u$. Lemma~\ref{prop1} and~\ref{prop2} verify the intuition that a random walk starting at any vertex $u$ will likely visit \emph{nearby} vertices first before visiting \emph{distant} vertices, and that a distant vertex $v$ is more reachable from some neighbor $j$ in less steps. This makes random walks more suitable for capturing communities rather global structural similarities among all the vertices.    
\begin{mylem} \label{prop3}
Suppose we start $d_u$ random walks of length $L$ from any vertex $u \in V$ in $G$. For a given edge $e = (v,v')$, let $I_e$ denote the total number of random walks containing $e$. Then, the expectation of the random variable $I_e$ is upper bounded by $L$, \ie, $\mathbb{E}[I_e] \leq L$.
\end{mylem}
\textsc{Proof.} Recall that the probability of a random walk starting at $u$ visits $v$ at time $t$ is {\small$P_{uv}^t = \boldsymbol\lambda_u \mP^t \boldsymbol\lambda_v^\intercal$\normalsize}, where $\boldsymbol\lambda_u$ is the indicator vector for vertex $u$, which equals $1$ in coordinate $u$ and $0$ otherwise. Then, for a given edge $e = (v,v')$, the probability that the random walk visits $v$ at time $t$ and $v'$ at time $t+1$ is {\small$\boldsymbol\lambda_u \mP^t \boldsymbol\lambda_v^\intercal / d_v$\normalsize} (since the transition probability from $v$ to $v'$ is $1/d_v$).
Suppose we start $d_u$ random walks of length $L$ from $u$, let $I_e$ denote the total number of random walks containing $e$, then the expectation of $I_e$ is the sum of the probabilities that there exists a random walk visiting $e = (v,v')$ as follows
{\small
\begin{align}
\mathbb{E}[I_e] &\leq \sum_{t=1}^{L} \sum_{u} d_u  \boldsymbol\lambda_u \mP^t \boldsymbol\lambda_v^\intercal / d_v \nonumber \\
&=  \sum_{t=1}^{L} \mathbf{1} \mD \mP^t \boldsymbol\lambda_v^\intercal / d_v  =  \sum_{t=1}^{L} \mathbf{1} \mD \boldsymbol\lambda_v^\intercal / d_v = \sum_{t=1}^{L} 1 = L \nonumber
\end{align}
\normalsize}\noindent
where $\mD$ is the degree matrix with the $i$th diagonal entry is the vertex degree $d_i$, $\mathbf{1}$ is the unit vector with all entries equal to $1$, and {\small$(\mathbf{1} \mD) \mP^t = \mathbf{1} \mD$\normalsize}. \eproof

\section{Experiments} 
\label{sec:exp}
\begin{figure}[h!]
\vspace{-2mm}
\centering
\includegraphics[width=0.4\linewidth]{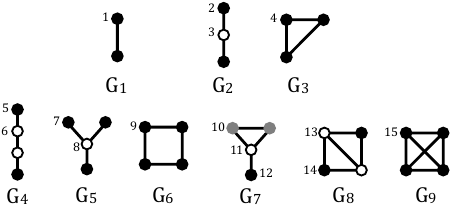}
\vspace{-2mm}
\caption{Summary of the 9 motifs and 15 orbits with 2-4 nodes.}
\label{fig:graphlet-attributes}
\vspace{-3mm}
\end{figure}

In this section, we investigate the effectiveness of the proposed framework using a variety of graphs.
Unless otherwise mentioned, all experiments use logarithmic binning\footnote{Logarithmic binning assigns the first $\delta N_v$ nodes with smallest attribute value to $0$ (where $0<\delta<1$), then assigns the $\delta$ fraction of remaining unassigned nodes with smallest value to $1$, and so on.}
and the bin size $\delta$ is chosen by searching over $\delta \in \{0.01, 0.1, 0.5, 0.9, 0.99\}$.
In these experiments, we use a simple function $\phi(\vx)$ that represents a concatenation of the attribute values in the node attribute vector $\vx$.
We searched over 10 subsets of the $9$ motif features of size 2-4 nodes shown in Figure~\ref{fig:graphlet-attributes}.
We evaluate the 
\emph{role2vec} approach presented in Section~\ref{sec:role2vec} that leverages the attributed random walk framework (Section~\ref{sec:framework}) against a number of popular methods including:
node2vec~\cite{node2vec},
DeepWalk~\cite{deepwalk}, 
struc2vec~\cite{struc2vec}, and 
LINE~\cite{line}.
For our approach and node2vec, we use the same hyperparameters ($\nD=128$, $\nR=10$, $\nL=80$) and grid search over $p,q\in \{0.25, 0.50, 1, 2, 4\}$ as mentioned in~\cite{node2vec}.
We use logistic regression (LR) with an L2 penalty.
The model is selected using 10-fold cross-validation on $10\%$ of the labeled data.
Experiments are repeated for 10 random seed initializations.
All results are statistically significant with p-value $< 0.01$.
We use AUC to evaluate the models.
Data was obtained from NetworkRepository~\cite{nr}.

\subsection{Comparison}
\label{sec:exp-comparison}
This section compares the proposed approach to other embedding methods for link prediction.
Given a partially observed graph $G$ with a fraction of missing edges, the link prediction task is to predict these missing edges.
We generate a labeled dataset of edges as done in~\cite{node2vec}.
Positive examples are obtained by removing $50\%$ of edges randomly, whereas \emph{negative examples} are generated by randomly sampling an equal number of node pairs that are not connected with an edge, \ie, each node pair $(i,j) \not\in E$. 
For each method, we learn features using the remaining graph that consists of only positive examples.
Using the learned embeddings from each method, we then learn a model to predict whether a given edge in the test set exists in $E$ or not.
Notice that node embedding methods such as $\deepwalk$ and node2vec require that each node in $G$ appear in at least one edge in the training graph (\ie, the graph remains connected), otherwise these methods are unable to derive features for such nodes.
This is a significant limitation that prohibits their use in many real-world applications.

\begin{figure}[h!]
\centering
\includegraphics[width=0.60\linewidth]{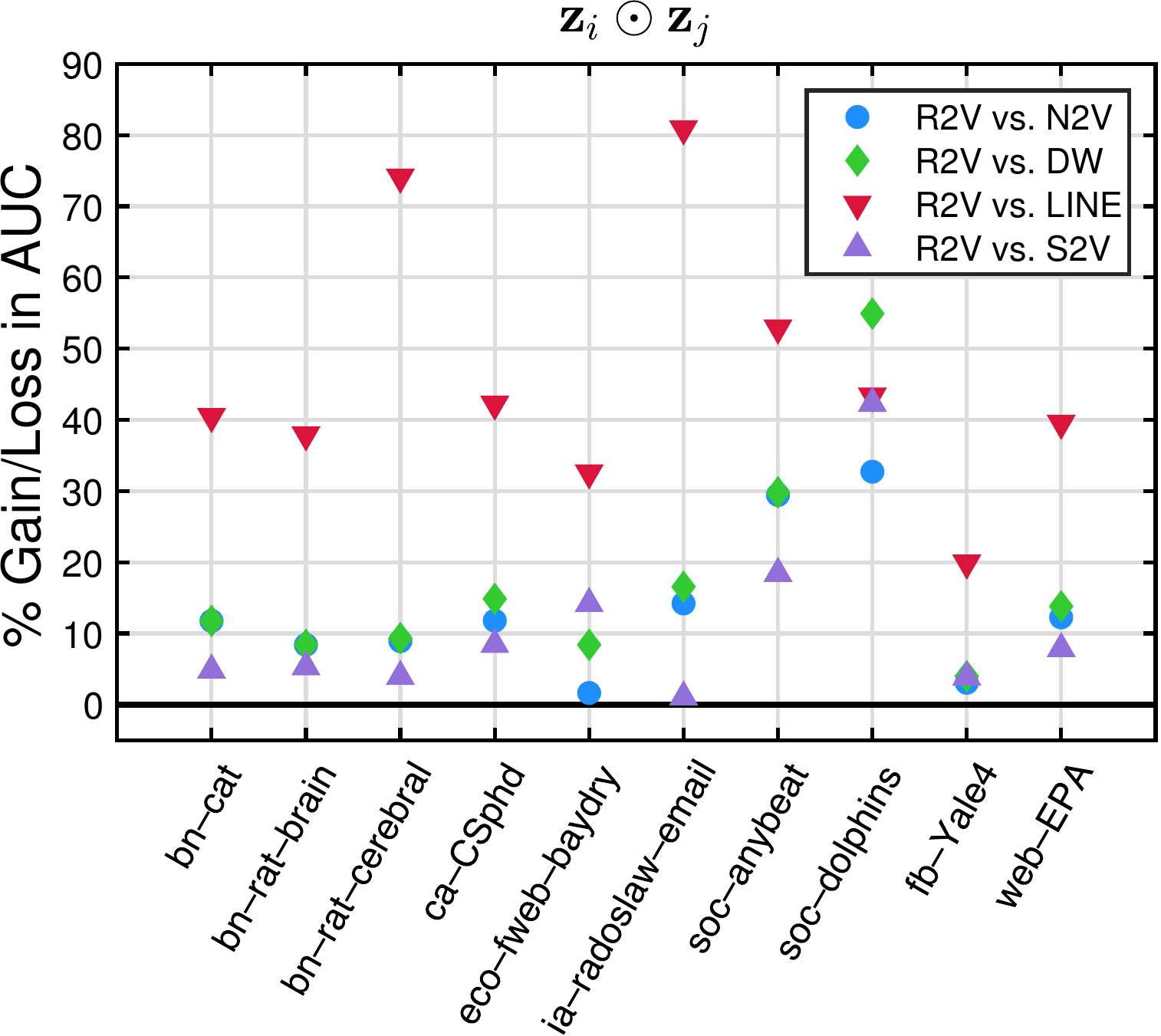} 
\vspace{-3mm}
\caption{AUC gain of Role2Vec (R2V) over the other methods for link prediction bootstrapped using Hadamard $\boldsymbol\alpha_i \odot \boldsymbol\alpha_j$.}
\label{fig:link-pred-perc-gain-auc-score-prod-op-rm}
\vspace{-3mm}
\end{figure}

\begin{table}[b!]
\centering
\setlength{\tabcolsep}{5.5pt}
\setlength{\tabcolsep}{2.5pt}
\ra{1.1}
\ra{1.05}
\scriptsize
\caption{AUC scores for various methods using $(\boldsymbol\alpha_i + \boldsymbol\alpha_j)\big/2$. Note \textbf{N2V}=node2vec, \textbf{DW}=$\deepwalk$ and \textbf{S2V}=struc2vec.}
\vspace{-2mm}
\label{table:link-pred-mean-OP}
\small
\fontsize{8.0}{9.0}\selectfont
\fontsize{7.5}{8.5}\selectfont
\begin{tabularx}{1.0\linewidth}{@{} r c Xc XXX X H}
\toprule
\textsc{Graph}  &&
\textbf{R2V} & 
\textbf{R2V-DW} &
\textbf{N2V} & 
\textbf{DW} & 
\textbf{LINE} &
\textbf{S2V} &
\\
\midrule

\dataName{bn\text{--}cat}  && 
\cellcolor{\verylightgreen} \text{$\mathbf{0.710}$}  &  
\cellcolor{\verylightgreen} \text{$0.688$}  &  
0.627  &  
0.627  &  
0.672  &
0.669 & 
\\

\dataName{bn\text{--}rat\text{--}brain}  && 
\cellcolor{\verylightgreen} \text{$\mathbf{0.748}$}  &  
0.731 & 
0.716  &  
0.716  &  
0.691  &  
0.729 & 
\\

\dataName{bn\text{--}rat\text{--}cerebral}  && 
\cellcolor{\verylightgreen} \text{$\mathbf{0.867}$}  &  
0.846 & 
0.813  &  
0.811  &  
0.709  &  
0.858 & 
\\

\dataName{ca\text{--}CSphd}  && 
\cellcolor{\verylightgreen} \text{$\mathbf{0.838}$}  &  
$\mathbf{0.838}$ & 
0.768  &  
0.735  &  
0.620  &  
0.791 & 
\\

\dataName{eco\text{--}fweb\text{--}baydry}  && 
\cellcolor{\verylightgreen} \text{$\mathbf{0.681}$}  &  
0.656 & 
0.655  &  
0.627  &  
0.660  &  
0.623 &  
\\

\dataName{ia\text{--}radoslaw\text{--}email}  && 
\cellcolor{\verylightgreen} \text{$\mathbf{0.867}$}  &  
0.847 & 
0.756  &  
0.745  &  
0.769  &  
0.857 & 
\\

\dataName{soc\text{--}anybeat}  && 
\cellcolor{\verylightgreen} \text{$\mathbf{0.961}$}  &  
0.960 & 
0.854  &  
0.848  &  
0.850  &  
0.883 & 
\\

\dataName{soc\text{--}dolphins}  && 
\cellcolor{\verylightgreen} \text{$\mathbf{0.656}$}  &  
0.597 & 
0.580  &  
0.498  &  
0.551  &  
0.590 & 
\\

\dataName{fb\text{--}Yale4}  && 
\cellcolor{\verylightgreen} \text{$\mathbf{0.793}$}  &  
$\mathbf{0.793}$ & 
0.742  &  
0.728  &  
0.763  &  
0.758 &  
\\

\dataName{web\text{--}EPA}  && 
\cellcolor{\verylightgreen} \text{$\mathbf{0.926}$}  &  
0.925 & 
0.804  &  
0.738  &  
0.768  &  
0.861 & 
\\

\bottomrule
\end{tabularx}
\end{table}

\begin{table}[h!]
\centering
\setlength{\tabcolsep}{2.5pt}
\ra{1.1}
\ra{1.05}
\scriptsize
\caption{
AUC scores for various methods using 
$\boldsymbol\alpha_i \odot \boldsymbol\alpha_j$.
Note \textbf{N2V}=node2vec, \textbf{DW}=$\deepwalk$ and \textbf{S2V}=struc2vec.
}
\vspace{-2mm}
\label{table:link-pred-prod-OP}
\small
\scriptsize
\fontsize{8.0}{9.0}\selectfont
\fontsize{7.5}{8.5}\selectfont
\begin{tabularx}{1.0\linewidth}{@{} r c Xc XXX X H}
\toprule
\textsc{Graph}  &&
\textbf{R2V} & 
\textbf{R2V-DW} &
\textbf{N2V} & 
\textbf{DW} & 
\textbf{LINE} &
\textbf{S2V} &
\\
\midrule

\dataName{bn\text{--}cat}  && 
\cellcolor{\verylightgreen} \text{$\mathbf{0.694}$}  &  
0.681 & 
0.621  &  
0.621  &  
0.494  &  
0.662 & 
\\

\dataName{bn\text{--}rat\text{--}brain}  && 
\cellcolor{\verylightgreen} \text{$\mathbf{0.775}$}  &  
$\mathbf{0.775}$ & 
0.715  &  
0.715  &  
0.562  &  
0.736 & 
\\

\dataName{bn\text{--}rat\text{--}cerebral}  && 
\cellcolor{\verylightgreen} \text{$\mathbf{0.867}$}  &  
0.838 & 
0.796  &  
0.793  &  
0.498  &  
0.834 &  
\\

\dataName{ca\text{--}CSphd}  && 
\cellcolor{\verylightgreen} \text{$\mathbf{0.758}$}  &  
0.738 & 
0.678  &  
0.660  &  
0.533  &  
0.699 & 
\\

\dataName{eco\text{--}fweb\text{--}baydry}  && 
\cellcolor{\verylightgreen} \text{$\mathbf{0.684}$}  &  
0.644 & 
0.673  &  
0.631  &  
0.516  &  
0.599 & 
\\

\dataName{ia\text{--}radoslaw\text{--}email}  && 
\cellcolor{\verylightgreen} \text{$\mathbf{0.852}$}  &  
0.821 & 
0.746  &  
0.731  &  
0.471  &  
0.843 & 
\\

\dataName{soc\text{--}anybeat}  && 
\cellcolor{\verylightgreen} \text{$\mathbf{0.945}$}  &   
$\mathbf{0.945}$ & 
0.730  &  
0.728  &  
0.618  &  
0.798 & 
\\

\dataName{soc\text{--}dolphins}  && 
\cellcolor{\verylightgreen} \text{$\mathbf{0.787}$}  &   
$\mathbf{0.787}$ & 
0.593  &  
0.508  &  
0.549  &  
0.553 & 
\\

\dataName{fb\text{--}Yale4}  && 
\cellcolor{\verylightgreen} \text{$\mathbf{0.940}$}  &  
0.906 & 
0.912  &  
0.904  &  
0.784  &  
0.905 & 
\\

\dataName{web\text{--}EPA}  && 
\cellcolor{\verylightgreen} \text{$\mathbf{0.907}$}  &  
0.885 & 
0.808  &  
0.797  &  
0.650  &  
0.841 & 
\\
\bottomrule
\end{tabularx}
\end{table}

For comparison, we use the same set of binary operators~\cite{node2vec} 
to construct features for the edges by combining the learned embeddings of its endpoints. 
The AUC results are provided in Table~\ref{table:link-pred-mean-OP} and~\ref{table:link-pred-prod-OP}.
Moreover, the AUC scores from $\ourMethod$ are all significantly better than the other methods at $p<0.01$. Note that we also used the \emph{role2vec} framework to generalize DeepWalk (DW) by using the notion of attributed random walk, we call this \textsc{R2V-DW}.  
We summarize the gain/loss in predictive performance over the other methods 
in Figure~\ref{fig:link-pred-perc-gain-auc-score-prod-op-rm}.
In all cases, $\ourMethod$ achieves better predictive performance over the other methods 
across a wide variety of graphs with different characteristics.
Overall, the mean and product binary operators give 
an average gain in predictive performance (over all graphs) of $11.1\%$ and $22\%$, respectively.

We also investigated learning types using low-rank matrix factorization (Eq.~\ref{eq:low-rank}) with squared loss.
No regularization or constraints were used.
Eq.~\ref{eq:k-means-objective-func} was used to partition nodes into $\m$ types.
Results are provided in Table~\ref{table:learning-types-low-rank} and are comparable to previous results that use concatenation to derive types.
Due to space, we report results for only a few graphs using the mean operator.

\begin{table}[h!]
\vspace{-2mm}
\centering
\fontsize{7.5}{8.5}\selectfont
\setlength{\tabcolsep}{5.5pt}
\scriptsize
\caption{AUC scores comparing types derived using concatenation vs. factorization.}
\label{table:learning-types-low-rank}
\vspace{-2mm}
\fontsize{7.5}{8.5}\selectfont
\begin{tabularx}{1.0\linewidth}{l H XXXX HHHHHH}
\toprule
&& 
\dataName{bn\text{--}cat} & 
\dataName{bn\text{--}rat\text{--}brain} & 
\dataName{ia\text{--}rado\text{--}email} &
\dataName{web\text{--}EPA}  &
\\
\midrule

\textbf{R2V} &&
\textbf{0.710} & 
0.748 & 
\textbf{0.867} & 
\textbf{0.926} & 
\\

\textbf{R2V-Fac.} 
&&
0.707 &
\textbf{0.761} &
0.848 &
0.905 &
\\

\bottomrule
\end{tabularx}
\end{table}

\subsection{Space-efficient Embeddings}
\label{sec:exp-space-efficiency}
We now investigate the space-efficiency of the learned embeddings from the proposed framework and intermediate representation.
Observe that any embedding method that implements the proposed attributed random walk framework (and intermediate representation) learns an embedding for each distinct node type $\y \in \Y$.
As described earlier in Sec.~\ref{sec:role2vec}, in the worst case, an embedding is learned for each of the $N_v$ vertices in the graph and we recover the baseline methods~\cite{deepwalk,node2vec} as a special case.
In general, the best embedding most often lies between such extremes and therefore the embedding learned from a method implementing \emph{Role2Vec} is often orders of magnitude smaller in size, since $\m \ll N_v$. 

\begin{figure}[h!]
\centering
\includegraphics[width=0.8\linewidth]{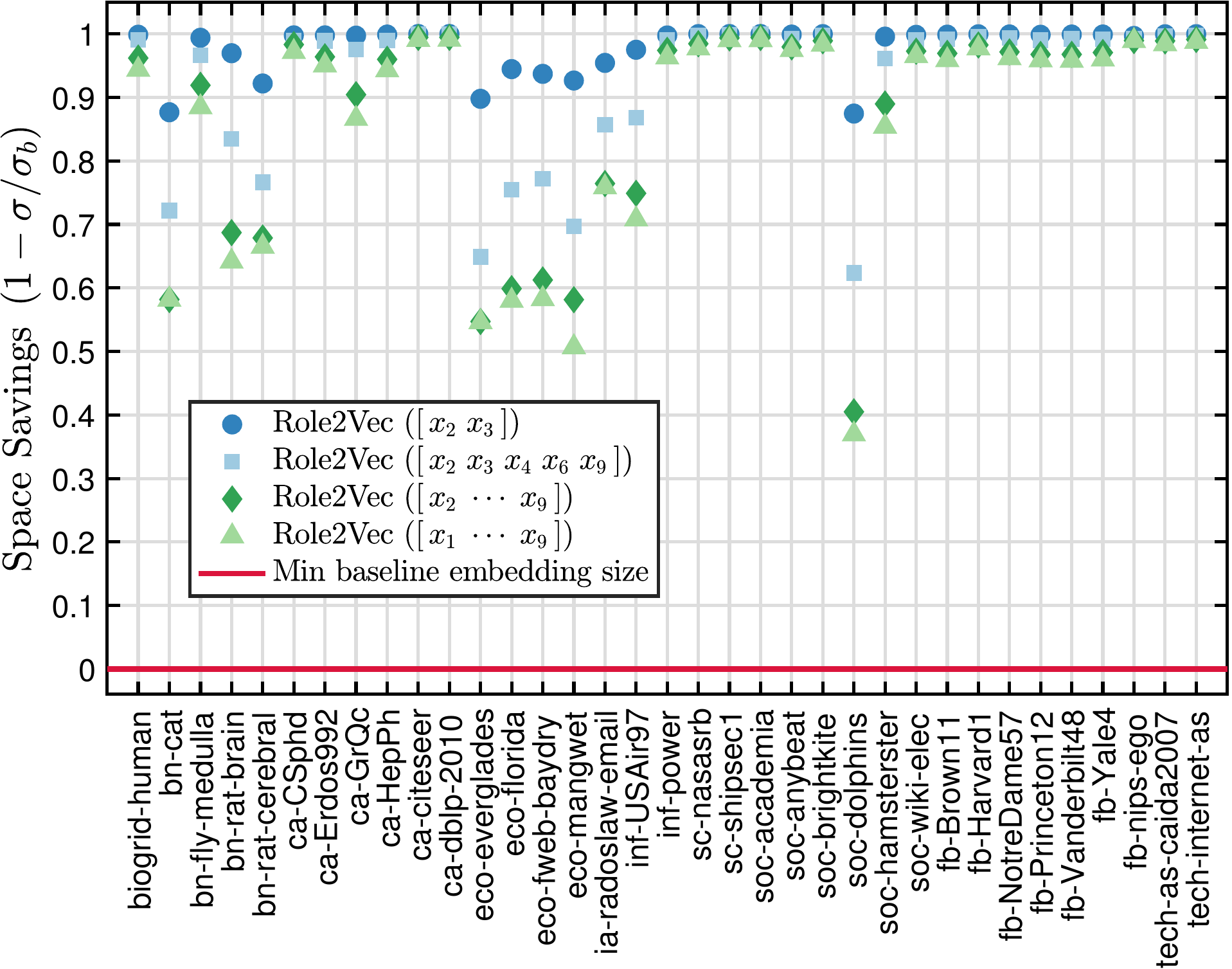} 
\vspace{-3mm}
\caption{
Evaluating the space savings of Role2Vec.
Space savings is defined as the reduction in embedding size (bytes) relative to the smallest embedding size among the baseline methods. 
Let $\sigma$ denote the size (bytes) of a Role2Vec embedding and $\sigma_{b}$ be the smallest embedding size from the baseline methods, then space savings is defined as $1 - \frac{\sigma}{\sigma_{b}}$.
Hence, larger values indicate a larger space savings (reduction in size).
Notably, the Role2Vec variants require significantly less space than existing methods as shown above.
Note $[\,\x_2 \; \x_3\,], \cdots, [\,\x_1\; \cdots \; \x_9\,]$ is the attribute sets in Eq.~\eqref{eq:space-1-2stars-triangles}-\eqref{eq:space-4-all} used as input to $\phi$ for mapping nodes to types.
}
\label{fig:space-efficient-embedding-reduction-gain}
\vspace{-3mm}
\end{figure}

Given an attribute vector $\vx$ of motif counts (Figure~\ref{fig:graphlet-attributes}) for an arbitrary node in $G$, we derive embeddings using each of the following:
{\fontsize{9}{10}\selectfont
\begin{align}
&\phi(\vx_i=[\,\x_2 \; \x_3\,]), \;\; \text{for } i=1,...,N_v \label{eq:space-1-2stars-triangles}\\
&\phi(\vx_i=[\,\x_2 \; \x_3\; \x_4\; \x_6\; \x_9\,]), \;\; \text{for } i=1,..., N_v \label{eq:space-2-2stars-triangles-x4-x6-4cliques} \\ 
&\phi(\vx_i=[\, \x_2 \; \x_3\; \x_4\; \x_5\, \cdots\, \x_9\,]), \;\; \text{for } i=1,..., N_v
\label{eq:space-3-all-but-degree} \\
&\phi(\vx_i=[\,\x_1\; \x_2 \; \x_3\; \x_4\; \x_5\, \cdots\, \x_9\,]), \;\; \text{for } i=1,...,N_v
\label{eq:space-4-all} 
\end{align}
}\normalsize
\noindent
where $\phi(\cdot)$ is a function that maps $\vx_i$ to a type $\y \in \Y$.
In these experiments, we use logarithmic binning (applied to each $N_v$-dimensional motif feature) with $\delta=0.5$ and use $\phi$ defined as the concatenation of the logarithmically binned attribute values.
Embeddings are learned using the different subsets of attributes in Eq.~\eqref{eq:space-1-2stars-triangles}-\eqref{eq:space-4-all}.
For instance, Eq.~\eqref{eq:space-1-2stars-triangles} indicates that vertex types are derived using the (logarithmic binned) number of 2-stars $\x_2$ and triangles $\x_3$ incident to the given vertex (Figure~\ref{fig:graphlet-attributes}).
We measure the space (in bytes) 
required to store the embedding learned by each method.
In Figure~\ref{fig:space-efficient-embedding-reduction-gain}, we summarize the reduction in space from our approach compared to the other methods.
In all cases, the embeddings learned from our approach require significantly less space and thus more space-efficient.
Specifically, the embeddings from our approach require on average $853$ times less space than the best method averaged across all graphs.

\section{Related Work} 
\label{sec:related-work}
\noindent
Recent embedding methods for graphs
have largely been based on the popular skip-gram model~\cite{mikolov2013distributed,cheng2006n} originally introduced for learning vector representations of words in text.
In particular, DeepWalk~\cite{deepwalk} used this approach to embed the nodes such that the co-occurrence frequencies of pairs in short random walks are preserved. 
Node2vec~\cite{node2vec} introduced hyperparameters to DeepWalk that tune the depth and breadth of the random walks.
These approaches are becoming increasingly popular and have been shown to outperform a number of existing methods.
These methods (and many others) are all based on simple random walks and thus are well-suited for generalization using the \emph{attributed random walk framework}.
\noindent
While most network representation learning methods use only the graph~\cite{deepwalk,line,grarep,node2vec}, our framework exploits both the graph \emph{and} structural features (\eg, motifs).

\noindent
While most work has focused on transductive (within-network) learning, there has been some recent work on graph-based inductive approaches.
Yang~\etal~\shortcite{Planetoid} proposed an inductive approach called Planetoid.
However, Planetoid is an embedding-based approach for semi-supervised learning and does not use any structural features.
Rossi~\etal~\shortcite{deepGL} proposed an inductive approach for (attributed) networks called DeepGL that learns (inductive) relational functions representing compositions of one or more operators applied to an initial set of graph features.
Recently, Hamilton~\etal~\shortcite{GraphSage} proposed a similar approach that also aggregates features from node neighborhoods.
However, these approaches are not based on random-walks.
Heterogeneous networks~\cite{shi2014hetesim} have also been recently considered~\cite{chang2015heterogeneous,dong2017metapath2vec} as well as attributed networks
~\cite{huang2017label,huang2017accelerated}.
Huang~\etal~\shortcite{huang2017label} proposed an approach for attributed networks with labels whereas Yang~\etal~\shortcite{yang2015network} used text features to learn node representations.
Liang~\etal~\shortcite{liang2017seano} proposed a semi-supervised approach for networks with outliers. 
Bojchevski~\etal~\shortcite{bojchevski2017deep} proposed an unsupervised rank-based approach.
Coley~\etal~\shortcite{coley2017convolutional} introduced a convolutional approach for attributed molecular graphs that learns graph embeddings as opposed to node embeddings. Duran~\etal~\cite{duran2017learning} proposed an embedding Propagation method to learn node representations.  
However,  most of these approaches are neither inductive nor space-efficient.

\section{Conclusion}
\label{sec:conc}
\vspace{-1mm}
This work proposed a flexible framework based on the notion of attributed random walks.
The framework serves as a basis for generalizing existing techniques (that are based on random walks) for use with attributed graphs, unseen nodes, graph-based transfer learning tasks, and allowing significantly larger graphs due to the inherent space-efficiency of the approach.
Instead of learning individual embeddings for each node, our approach learns embeddings for each type based on functions that map feature vectors to types.
This allows for both inductive and transductive learning.

\clearpage
\begin{small}
\bibliographystyle{named}
\bibliography{paper}

\begin{thebibliography}{}

\bibitem[\protect\citeauthoryear{Ahmed \bgroup \em et al.\egroup }{2015}]{pgd}
Nesreen~K. Ahmed, Jennifer Neville, Ryan~A. Rossi, and Nick Duffield.
\newblock Efficient graphlet counting for large networks.
\newblock In {\em ICDM}, page~10, 2015.

\bibitem[\protect\citeauthoryear{Ahmed \bgroup \em et al.\egroup
  }{2016}]{ahmed2016kais}
Nesreen~K. Ahmed, Jennifer Neville, Ryan~A. Rossi, Nick Duffield, and
  Theodore~L. Willke.
\newblock Graphlet decomposition: Framework, algorithms, and applications.
\newblock {\em KAIS}, pages 1--32, 2016.

\bibitem[\protect\citeauthoryear{Akoglu \bgroup \em et al.\egroup
  }{2015}]{akoglu2015graph}
Leman Akoglu, Hanghang Tong, and Danai Koutra.
\newblock Graph based anomaly detection and description: a survey.
\newblock {\em DMKD}, 29(3):626--688, 2015.

\bibitem[\protect\citeauthoryear{Al~Hasan and Zaki}{2011}]{al2011survey}
Mohammad Al~Hasan and Mohammed~J Zaki.
\newblock A survey of link prediction in social networks.
\newblock In {\em Social Network Data Analytics}, pages 243--275. 2011.

\bibitem[\protect\citeauthoryear{Benson \bgroup \em et al.\egroup
  }{2016}]{benson2016higher}
Austin~R Benson, David~F Gleich, and Jure Leskovec.
\newblock Higher-order organization of complex networks.
\newblock {\em Science}, 353(6295):163--166, 2016.

\bibitem[\protect\citeauthoryear{Bojchevski and
  G{\"u}nnemann}{2017}]{bojchevski2017deep}
Aleksandar Bojchevski and Stephan G{\"u}nnemann.
\newblock Deep gaussian embedding of attributed graphs: Unsupervised inductive
  learning via ranking.
\newblock {\em arXiv:1707.03815}, 2017.

\bibitem[\protect\citeauthoryear{Cao \bgroup \em et al.\egroup }{2015}]{grarep}
Shaosheng Cao, Wei Lu, and Qiongkai Xu.
\newblock Grarep: Learning graph representations with global structural
  information.
\newblock In {\em CIKM}, pages 891--900. ACM, 2015.

\bibitem[\protect\citeauthoryear{Cavallari \bgroup \em et al.\egroup
  }{2017}]{ComE}
Sandro Cavallari, Vincent~W Zheng, Hongyun Cai, Kevin Chen-Chuan Chang, and
  Erik Cambria.
\newblock Learning community embedding with community detection and node
  embedding on graphs.
\newblock In {\em CIKM}, pages 377--386, 2017.

\bibitem[\protect\citeauthoryear{Chang \bgroup \em et al.\egroup
  }{2015}]{chang2015heterogeneous}
Shiyu Chang, Wei Han, Jiliang Tang, Guo-Jun Qi, Charu~C Aggarwal, and Thomas~S
  Huang.
\newblock Heterogeneous network embedding via deep architectures.
\newblock In {\em SIGKDD}, pages 119--128, 2015.

\bibitem[\protect\citeauthoryear{Cheng \bgroup \em et al.\egroup
  }{2006}]{cheng2006n}
Winnie Cheng, Chris Greaves, and Martin Warren.
\newblock From n-gram to skipgram to concgram.
\newblock {\em Int. J. of Corp. Linguistics}, 11(4):411--433, 2006.

\bibitem[\protect\citeauthoryear{Coley \bgroup \em et al.\egroup
  }{2017}]{coley2017convolutional}
Connor~W Coley, Regina Barzilay, William~H Green, Tommi~S Jaakkola, and Klavs~F
  Jensen.
\newblock Convolutional embedding of attributed molecular graphs for physical
  property prediction.
\newblock {\em J. Chem. Info. \& Mod.}, 2017.

\bibitem[\protect\citeauthoryear{Dong \bgroup \em et al.\egroup
  }{2017}]{dong2017metapath2vec}
Yuxiao Dong, Nitesh~V Chawla, and Ananthram Swami.
\newblock metapath2vec: Scalable representation learning for heterogeneous
  networks.
\newblock In {\em SIGKDD}, pages 135--144, 2017.

\bibitem[\protect\citeauthoryear{Duran and Niepert}{2017}]{duran2017learning}
Alberto~Garcia Duran and Mathias Niepert.
\newblock Learning graph representations with embedding propagation.
\newblock In {\em NIPS}, pages 5119--5130, 2017.

\bibitem[\protect\citeauthoryear{Getoor and Taskar}{2007}]{introSRL07}
L.~Getoor and B.~Taskar, editors.
\newblock {\em Intro. to SRL}.
\newblock MIT Press, 2007.

\bibitem[\protect\citeauthoryear{Goldsmith and
  Davenport}{1990}]{goldsmith1990assessing}
Timothy~E Goldsmith and Daniel~M Davenport.
\newblock Assessing structural similarity of graphs.
\newblock 1990.

\bibitem[\protect\citeauthoryear{Goyal and Ferrara}{2017}]{goyal2017graph}
Palash Goyal and Emilio Ferrara.
\newblock Graph embedding techniques, applications, and performance: A survey.
\newblock {\em arXiv preprint arXiv:1705.02801}, 2017.

\bibitem[\protect\citeauthoryear{Grover and Leskovec}{2016}]{node2vec}
Aditya Grover and Jure Leskovec.
\newblock node2vec: Scalable feature learning for networks.
\newblock In {\em SIGKDD}, pages 855--864, 2016.

\bibitem[\protect\citeauthoryear{Hamilton \bgroup \em et al.\egroup
  }{2017}]{GraphSage}
William~L Hamilton, Rex Ying, and Jure Leskovec.
\newblock Inductive representation learning on large graphs.
\newblock {\em arXiv:1706.02216}, 2017.

\bibitem[\protect\citeauthoryear{Harris}{1954}]{harris1954distributional}
Zellig~S Harris.
\newblock Distributional structure.
\newblock {\em Word}, 10(2-3):146--162, 1954.

\bibitem[\protect\citeauthoryear{Henderson \bgroup \em et al.\egroup
  }{2011}]{henderson2011s}
Keith Henderson, Brian Gallagher, Lei Li, Leman Akoglu, Tina Eliassi-Rad,
  Hanghang Tong, and Christos Faloutsos.
\newblock It's who you know: graph mining using recursive structural features.
\newblock In {\em SIGKDD}, pages 663--671. ACM, 2011.

\bibitem[\protect\citeauthoryear{Huang \bgroup \em et al.\egroup
  }{2017a}]{huang2017accelerated}
Xiao Huang, Jundong Li, and Xia Hu.
\newblock Accelerated attributed network embedding.
\newblock In {\em SDM}, 2017.

\bibitem[\protect\citeauthoryear{Huang \bgroup \em et al.\egroup
  }{2017b}]{huang2017label}
Xiao Huang, Jundong Li, and Xia Hu.
\newblock Label informed attributed network embedding.
\newblock In {\em WSDM}, 2017.

\bibitem[\protect\citeauthoryear{Koyut{\"u}rk \bgroup \em et al.\egroup
  }{2006}]{koyuturk2006pairwise}
Mehmet Koyut{\"u}rk, Yohan Kim, Umut Topkara, Shankar Subramaniam, Wojciech
  Szpankowski, and Ananth Grama.
\newblock Pairwise alignment of protein interaction networks.
\newblock {\em JCB}, 13(2):182--199, 2006.

\bibitem[\protect\citeauthoryear{Kuwadekar and
  Neville}{2011}]{kuwadekar2011relational}
Ankit Kuwadekar and Jennifer Neville.
\newblock Relational active learning for joint collective classification
  models.
\newblock In {\em ICML}, pages 385--392, 2011.

\bibitem[\protect\citeauthoryear{Liang \bgroup \em et al.\egroup
  }{2017}]{liang2017seano}
Jiongqian Liang, Peter Jacobs, and Srinivasan Parthasarathy.
\newblock Seano: Semi-supervised embedding in attributed networks with
  outliers.
\newblock {\em arXiv:1703.08100}, 2017.

\bibitem[\protect\citeauthoryear{Lorrain and
  White}{1977}]{lorrain1977structural}
Francois Lorrain and Harrison~C White.
\newblock Structural equivalence of individuals in social networks.
\newblock In {\em Social Networks}, pages 67--98. Elsevier, 1977.

\bibitem[\protect\citeauthoryear{Mikolov \bgroup \em et al.\egroup
  }{2013}]{mikolov2013distributed}
Tomas Mikolov, Ilya Sutskever, Kai Chen, Greg~S Corrado, and Jeff Dean.
\newblock Distributed representations of words and phrases and their
  compositionality.
\newblock In {\em Advances in neural information processing systems}, pages
  3111--3119, 2013.

\bibitem[\protect\citeauthoryear{Neville and
  Jensen}{2000}]{neville2000iterative}
Jennifer Neville and David Jensen.
\newblock Iterative classification in relational data.
\newblock In {\em AAAI SRL Workshop}, pages 13--20, 2000.

\bibitem[\protect\citeauthoryear{Ng \bgroup \em et al.\egroup
  }{2002}]{ng2002spectral}
Andrew~Y Ng, Michael~I Jordan, and Yair Weiss.
\newblock On spectral clustering: Analysis and an algorithm.
\newblock In {\em NIPS}, pages 849--856, 2002.

\bibitem[\protect\citeauthoryear{Perozzi \bgroup \em et al.\egroup
  }{2014}]{deepwalk}
Bryan Perozzi, Rami Al-Rfou, and Steven Skiena.
\newblock Deepwalk: Online learning of social representations.
\newblock In {\em SIGKDD}, pages 701--710, 2014.

\bibitem[\protect\citeauthoryear{Ribeiro \bgroup \em et al.\egroup
  }{2017}]{struc2vec}
Leonardo~F.R. Ribeiro, Pedro~H.P. Saverese, and Daniel~R. Figueiredo.
\newblock Struc2vec: Learning node representations from structural identity.
\newblock In {\em SIGKDD}, pages 385--394, 2017.

\bibitem[\protect\citeauthoryear{Rossi and Ahmed}{2015a}]{rossi2014roles}
R.A. Rossi and N.K. Ahmed.
\newblock Role discovery in networks.
\newblock {\em TKDE}, 27(4):1112--1131, 2015.

\bibitem[\protect\citeauthoryear{Rossi and Ahmed}{2015b}]{nr}
Ryan~A. Rossi and Nesreen~K. Ahmed.
\newblock The network data repository with interactive graph analytics and
  visualization.
\newblock In {\em AAAI}, pages 4292--4293, 2015.

\bibitem[\protect\citeauthoryear{Rossi \bgroup \em et al.\egroup
  }{2017}]{deepGL}
Ryan~A. Rossi, Rong Zhou, and Nesreen~K. Ahmed.
\newblock Deep feature learning for graphs.
\newblock In {\em arXiv:1704.08829}, 2017.

\bibitem[\protect\citeauthoryear{Rudolph \bgroup \em et al.\egroup
  }{2016}]{rudolph2016exponential}
Maja Rudolph, Francisco Ruiz, Stephan Mandt, and David Blei.
\newblock Exponential family embeddings.
\newblock In {\em Advances in Neural Information Processing Systems}, pages
  478--486, 2016.

\bibitem[\protect\citeauthoryear{Shi \bgroup \em et al.\egroup
  }{2014}]{shi2014hetesim}
Chuan Shi, Xiangnan Kong, Yue Huang, S~Yu Philip, and Bin Wu.
\newblock {HeteSim: A General Framework for Relevance Measure in Heterogeneous
  Networks}.
\newblock {\em TKDE}, 26(10):2479--2492, 2014.

\bibitem[\protect\citeauthoryear{Tang \bgroup \em et al.\egroup }{2015}]{line}
Jian Tang, Meng Qu, Mingzhe Wang, Ming Zhang, Jun Yan, and Qiaozhu Mei.
\newblock {LINE: Large-scale Information Network Embedding}.
\newblock In {\em WWW}, pages 1067--1077, 2015.

\bibitem[\protect\citeauthoryear{Yang \bgroup \em et al.\egroup
  }{2015}]{yang2015network}
Cheng Yang, Zhiyuan Liu, Deli Zhao, Maosong Sun, and Edward~Y Chang.
\newblock Network representation learning with rich text information.
\newblock In {\em IJCAI}, pages 2111--2117, 2015.

\bibitem[\protect\citeauthoryear{Yang \bgroup \em et al.\egroup
  }{2016}]{Planetoid}
Zhilin Yang, William~W Cohen, and Ruslan Salakhutdinov.
\newblock Revisiting semi-supervised learning with graph embeddings.
\newblock {\em arXiv:1603.08861}, 2016.

\bibitem[\protect\citeauthoryear{Zager and Verghese}{2008}]{zager2008graph}
Laura~A Zager and George~C Verghese.
\newblock Graph similarity scoring and matching.
\newblock {\em Applied mathematics letters}, 21(1):86--94, 2008.

\end{thebibliography}
\end{small}

\end{document}